%% file: main.tex
\documentclass[10pt,twocolumn,letterpaper]{article}

\usepackage[pagenumbers]{cvpr} % To force page numbers, e.g. for an arXiv version
\usepackage{pifont}
\usepackage{adjustbox}
\usepackage{multirow}
\usepackage{cuted}
\usepackage{float}
\usepackage[linesnumbered,ruled,vlined]{algorithm2e}
\usepackage{fancyvrb}
\usepackage{stfloats}
\usepackage{soul}
\usepackage{enumitem}
\usepackage{tikz}
\usepackage{graphicx}
\usepackage{pgfplots}
\usepackage{enumitem}
\usepackage{amsmath}
\usepackage{amssymb}
\usepackage{mathtools}
\usepackage{amsthm}
\usepackage{paralist}
\newcommand*\circled[1]{\tikz[baseline=(char.base)]{
            \node[shape=circle,draw,inner sep=.4pt] (char) {#1};}}
\usepackage[flushleft, para]{threeparttable}
\usepackage[operators,sets]{cryptocode}


\definecolor{cvprblue}{rgb}{0.21,0.49,0.74}
\usepackage[pagebackref,breaklinks,colorlinks,allcolors=cvprblue]{hyperref}

\def\paperID{9} % *** Enter the Paper ID here
\def\confName{CVPR}
\def\confYear{2025}

\title{Provenance Detection for AI-Generated Images: Combining Perceptual Hashing, Homomorphic Encryption, and AI Detection Models}

\author{
   Shree Singhi\textsuperscript{1,2$\star$} \ \ \
   Aayan Yadav\textsuperscript{1,2$\star$} \ \ \
   Aayush Gupta\textsuperscript{3} \ \ \
   Shariar Ebrahimi\textsuperscript{4} \ \ \
   Parisa Hassanizadeh\textsuperscript{5} \\
   \textsuperscript{2}Zellic\ \ \
   \textsuperscript{3}MIT, ZK Email\ \ \
   \textsuperscript{4}Newcastle University \ \ \
   \textsuperscript{5}Polish Academy of Science \\
   \small \url{https://proteus.photos}
}

\begin{document}

 \maketitle

 \begin{strip}
    \centering
    \includegraphics[width=\textwidth]{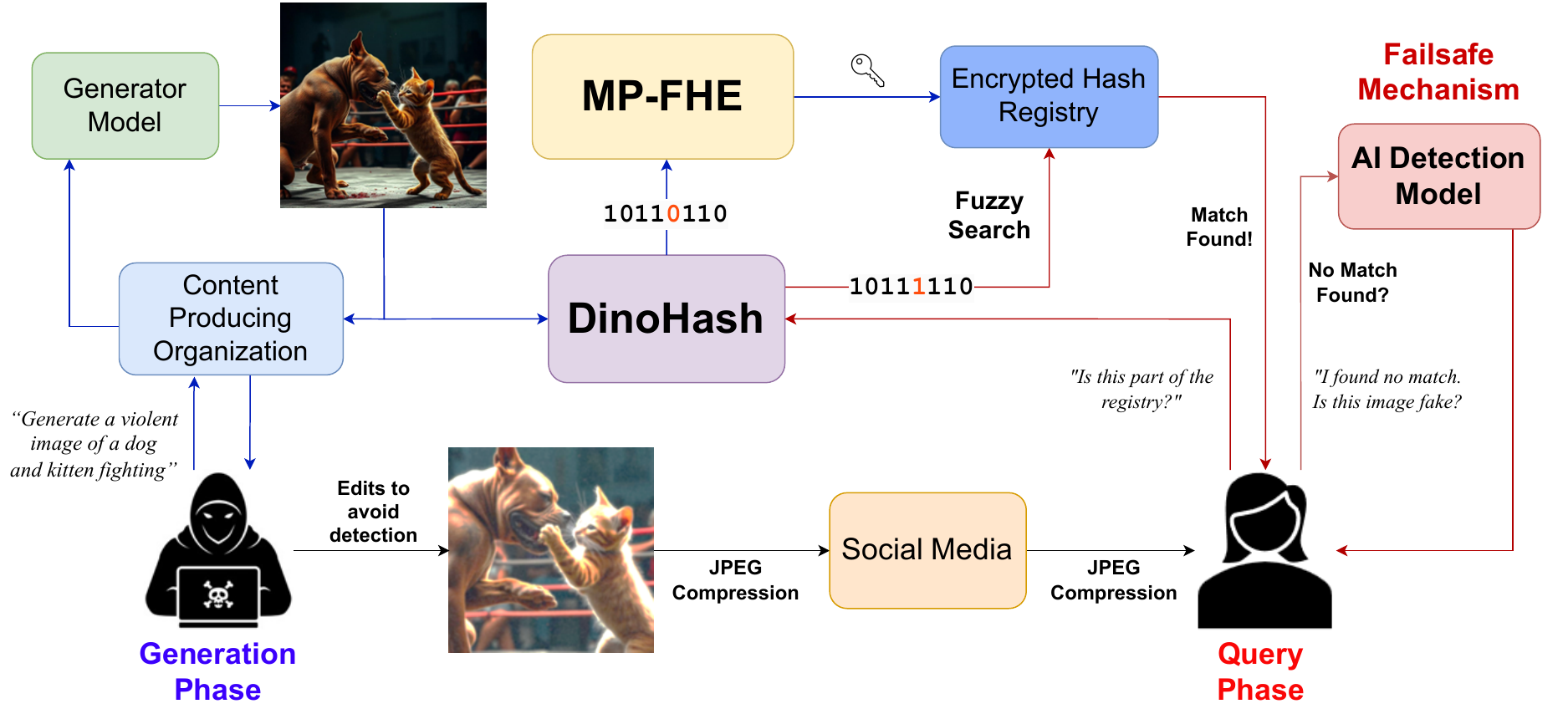}
    \captionof{figure}{\textbf{System Overview.} We combine the strengths of our own perceptual hashing model, DinoHash, Multi-Party Fully Homomorphic Encryption (MP-FHE) and improved AI Detection Model to build a robust provenance framework.}
    \label{fig:teaser}
\end{strip}

\input{sec/0_abstract}
\input{sec/1_intro}

\input{sec/2_background}

\input{sec/3_motivation}
\input{sec/4_proposed}

\input{sec/5_privacy_preserving}
\input{sec/6_conclusion}
{
    \small
    \bibliographystyle{ieeenat_fullname}
    \bibliography{main}
}

\input{sec/appendix}

\end{document}

%% file: sec/0_abstract.tex
\begin{abstract}

As AI-generated sensitive images become more prevalent, identifying their source is crucial for distinguishing them from real images. Conventional image watermarking methods are vulnerable to common transformations like filters, lossy compression, and screenshots, often applied during social media sharing. Watermarks can also be faked or removed if models are open-sourced or leaked since images can be rewatermarked. We have developed a three-part framework for secure, transformation-resilient AI content provenance detection, to address these limitations. We develop an adversarially robust state-of-the-art perceptual hashing model, DinoHash, derived from DINOV2, which is robust to common transformations like filters, compression, and crops. Additionally, we integrate a Multi-Party Fully Homomorphic Encryption~(MP-FHE) scheme into our proposed framework to ensure the protection of both user queries and registry privacy. Furthermore, we improve previous work on AI-generated media detection. This approach is useful in cases where the content is absent from our registry. DinoHash significantly improves average bit accuracy by $12\%$ over state-of-the-art watermarking and perceptual hashing methods while maintaining superior true positive rate (TPR) and false positive rate (FPR) tradeoffs across various transformations. Our AI-generated media detection results show a $25\%$ improvement in classification accuracy on commonly used real-world AI image generators over existing algorithms. By combining perceptual hashing, MP-FHE, and an AI content detection model, our proposed framework provides better robustness and privacy compared to previous work.
\clearpage

% In a perfect world, you might imagine that all content on the web is marked with some kind of invisible identifier. The identifier could contain a cryptographic signature attesting to the origin of the content, or additional metadata like the model used to create a deepfake or the approximate location of a photo. The ideal identifier should be imperceptible to human senses and impervious to algorithmic analysis. It should also remain secure against both unintentional attacks like compression, screenshotting, and re-capture, and intentional attacks, such as histogram or Gaussian noise manipulation. Unfortunately, watermarking schemes do not have proper robustness against “realistic” internet interactions such as JPEG compression, a non-corrupting crop, or screenshots. \\

% Embedding-derived perceptual hashes allow us to track a content’s perceptual representation in a quick and robust way, giving us a “weak” mapping from an image to similar images. In addition, they are more robust against aforementioned attacks, compared to watermarking. Finally, they can be compared in a completely privacy preserving manner. \\

% We introduce a state-of-the-art perceptual hash (p-hash) algorithm for image similarity, as well as a novel system for indexing images using p-hash values and verifying content originality through MPC queries. To ensure privacy, the p-hash values are stored encrypted in an external database. Authorities such as content providers can use a shared public key to query whether query images are already in the database, or are unique.

\end{abstract}

%% file: sec/1_intro.tex
\vspace{-40pt}

\section{Introduction}
\label{sec:intro}

Ensuring the authenticity and provenance of digital visual content is increasingly critical as AI-generated media becomes more prevalent. Traditional watermarking methods have limited robustness to everyday transformations and raise security concerns if models are distributed for local use. C2PA~\cite{c2pa2023} does not work in practical scenarios, due to the signature becoming invalid upon any image transformation. Proof of transformation helps maintain validation after image edits~\cite{vimz, veritas}, but we do not expect many image editors to integrate this technology. To tackle these challenges, we introduce a three-part framework aimed at secure and reliable content provenance detection for AI-generated content in real world.

First, we introduce a perceptual hashing algorithm based on DINOV2 \cite{oquab2023dinov2}, a feature extraction network, that captures the semantic and structural features of images, providing an alternative to watermarking by remaining resilient to common image transformations. Second, to secure the integrity of the provenance detection process, we employ Mutli-Party Fully Homomorphic Encryption~(MP-FHE)~\cite{mouchet2021multiparty}, which allows operations on encrypted data, enabling secure, privacy-preserving computations without compromising the system's security regarding data leakage \cite{gentry2009fully}. Lastly, we extend prior work in AI-generated content detection by training a state of the art model to identify and distinguish AI-generated images directly, in case images are not found in our database.

Deep watermarking algorithms embed a nearly invisible hash into an image, which can later be extracted using a specialized network to verify and match images~\cite{deep_watermark}. The main objectives of watermarking are~(a)~robustness against natural image distortions and~(b)~minimal alteration of the original image. Most deep watermarking papers report remarkable results on simple spatial distortions such as brightness or contrast, cropping, and flipping. However, they often fail when exposed to destructive frequency-domain transformations, including blurring, JPEG compression, screenshots, and other non-differentiable processes. This limitation arises because watermarking methods embed information in an image’s high-frequency components, which are often degraded via compression during everyday manipulations like social media uploads~\cite{laghari2018assessment}, sharing on messaging apps~\cite{anwar2021image}, format conversions, resizing, and embedding in videos~\cite{YoutubeCompression}, making watermarking ineffective for content provenance detection in practical settings. 

Traditionally, watermarking requires training a separate deep-learning model to embed watermarks into an image, making them easily removable. However, Meta's recently developed framework, Stable Signature~\cite{meta2023stablesig}, integrates watermarking directly into the diffusion model through fine-tuning, eliminating the need for an additional model during inference. While this approach reduces inference costs, it introduces the burden of maintaining a unique set of model weights for each user. Given that the number of customers of diffusion models is in the millions, this storage requirement renders the approach impractical for large-scale industry deployment unless the model is provided to users for local inference. However, as the authors demonstrated, releasing the model to users introduces vulnerabilities due to aversion techniques like model purification and collusion, undermining the security and purpose of the watermarking system.

Additionally, watermarking methods might introduce unwanted visible artifacts \cite{waves}. In contrast, perceptual hashes offer a potential alternative. Instead of embedding data, perceptual hashing derives a hash from the image’s semantic embeddings and structural features, which remain largely unaffected by common transformations, making it more suitable for content provenance applications \cite{phash2020}.

%-------------------------------------------------------------------------
After generating perceptual hashes, it's crucial to query them privately to protect sensitive information. For example, closed-source content generators~(such as OpenAI or Canon) should be able to store perceptual hashes of user-generated images in a public database without leaking information that could lead to prompt reconstruction. Likewise, users need to query the database without revealing all their images to the content provider. Several systems have been developed to address this private approximate nearest neighbors search~(ANNS/PNNS) challenge, including Janus~\cite{edalatnejad2024janus}, Worldcoin's Iris~\cite{iris-search} matching system, Apple's Wally~\cite{wally-search}, and Panther~\cite{li2024panther}.
The Worldcoin Iris Matching system uses secure multiparty computation~(MPC) to perform privacy-preserving fuzzy matching of iris hash codes, which are similar to our proposed perceptual hashes. It operates under a non-collusion assumption between servers to ensure privacy. In contrast, Apple's Wally system introduces differential privacy by adding fake queries to hide the original query, and querying clusters of data rather than the entire database, thus trading off some accuracy for enhanced performance. Wally uses Somewhat Homomorphic Encryption~(SHE) for secure lookups, based on the BFV~\cite{fan2012somewhat, brakerski2012fully} and BGV~\cite{brakerski2014leveled} schemes, which allow a single entity to decrypt any ciphertext without restriction. In contrast, our system is built on a boolean FHE scheme, Multi-party FHEW~(MP-FHE), which requires multiple parties to participate in decryption~\cite{mouchet2021multiparty, lee2023efficient}. This enables the design of PNNS systems with stronger security guarantees, ensuring that no single entity can access plaintext data without the consent of all other parties.
\begin{figure*}[!t]
    \centering
    \clearpage
    \includegraphics[page=2, width=0.9\linewidth]{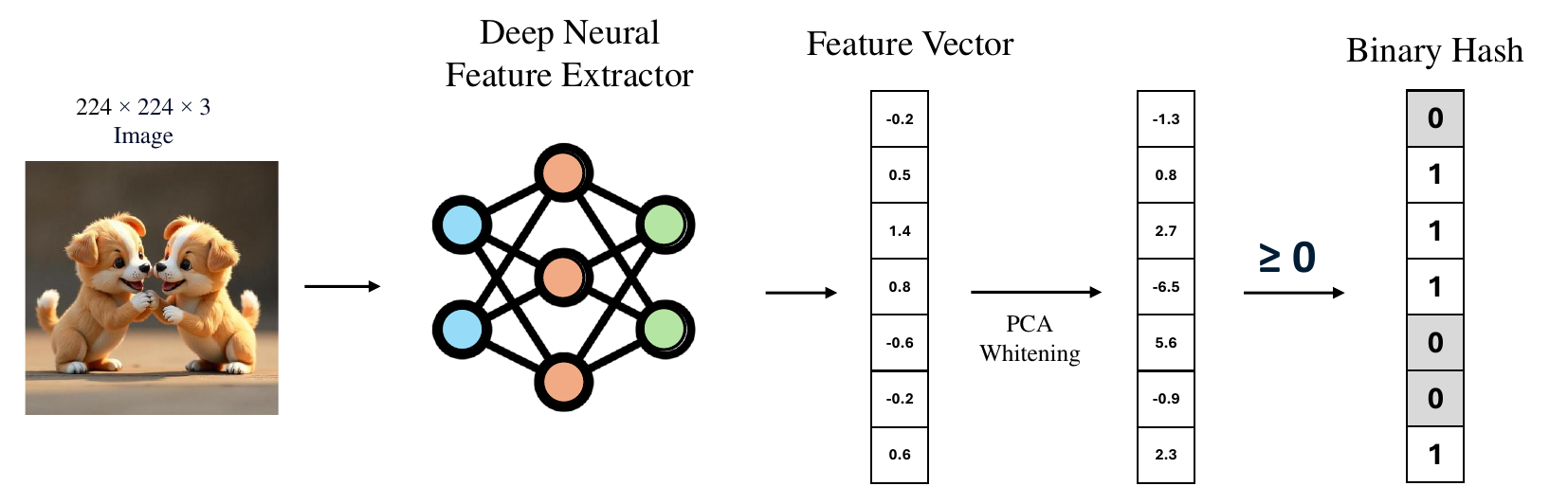}
    \caption{\textbf{Deep neural feature extractor architecture}. The pipeline consists of a feature extractor network and Locality Sensitive Hashing~(LSH) step. The PCA whitening decorrelates components of the feature vector, so that the hashes of two random different images have the least probability of being the same.}
    \label{fig:perceptual_hash}
\end{figure*}
Modern tools and open-source models have democratized access to synthetic image generation. In this reality, it is unreasonable to expect every image appearing on the internet to be hashed and stored in a registry by some content-producing organization~(CPO) like cameras or AI image APIs. So, we build a purely deep learning-based detector to perform the binary classification task of detection of synthetic images.

\noindent In conclusion, we have four main contributions:
\begin{compactenum}
    \item A deep perceptual hashing algorithm that is robust to a wide range of image transformations.
    \item A cryptographic matching system that privately looks up the provenance of a perceptual hash in a registry, without leaking any data about the registry items or the queried perceptual hash.
    \item A deep learning based detector to detect synthetic images that are not stored in the registry.
    \item A novel approach to fine-tune any deep-learning-based perceptual hashing model to be adversarially robust. Although several works explore adversarial attacks against perceptual hashing models, to the best of our knowledge, we are the first to introduce an adversarial training mechanism for them.
\end{compactenum}

%% file: sec/2_background.tex
\section{Background}
In this section, we discuss related work and the essential concepts used our framework. We provide an overview of perceptual hashing techniques, MPC methods, and deep learning-based detectors.

\subsection{Perceptual Hashing}

Since the MP-FHE scheme requires fuzzy hashes, we require a hashing algorithm that is mostly invariant to perceptual transformations of the image. Several perceptual hashing algorithms exist including aHash (Average Hashing), mHash (Median Hashing), dHash (Difference Hashing), bHash (Block Hashing), wHash (Wavelet Hashing) and the Discrete Fourier Transform Perceptual Hash \cite{Zauner2010ImplementationAB}. These hashing schemes are used by services like Microsoft’s PhotoDNA and Facebook’s PDQ. These hashing algorithms work by dividing images into squares, converting them into a black-and-white format, and quantifying the shading of the squares in different ways. Although these hashes are mostly invariant to color manipulation and noise, they fail when the images are subjected to crops. This is because cropping the image offsets the alignment of the square blocks, heavily altering the hash. 

In recent years, numerous deep hashing algorithms based on convolutional neural networks have been developed~\cite{deep_hashing_compact_codes, deep_supervised_hashing_fast_retrieval, deep_semantic_ranking_hashing_multilabel, deep_supervised_hashing_large_scale}. These methods rely on deep neural networks to extract image features and subsequently compute a hash value based on those features. The resulting hashes capture the semantic content of the image and exhibit greater robustness to transformations like cropping and color adjustments. Apple's Neuralhash \cite{struppek2022learning} was developed for Child Sexual Abuse Material (CSAM) detection. \autoref{fig:perceptual_hash} illustrates the high level architecture of our chosen deep hashing algorithm, which we will now discuss in more detail.

A perceptual hash function $H \colon \mathbb{R}^{H \times W \times C} \to \{0,1\}^k$ maps an image $x$ to a $k$-bit binary hash. Let $H(x)$ denote the hash function, where $H(x) = \bigl( h_1(x), \ldots, h_k(x) \bigr)$. These algorithms consist of two main components. First, a deep feature extractor $D(x)\colon \mathbb{R}^{H \times W \times C} \to \mathbb{R}^{k}$ generates a feature vector $z \in \mathbb{R}^{k}$ from the image $x$. This feature vector numerically represents the semantic and structural content of the image. Next, locality-sensitive hashing (LSH)~\cite{locality, jafari2021surveylocalitysensitivehashing} is used to assign similar feature vectors to buckets with similar hash values. For each of the $k$ features, the bit value $h_i(x)$ is set to either $1$ or $0$ depending on the sign of $z_i$ using the heaviside step function.

The idea is that transformations applied to the original image do not perturb the feature vector and, consequently, the hash significantly. We would like it if each $h_i(x)$ follows an i.i.d Bernoulli distribution with parameter $0.5$. This is achieved by applying PCA whitening to the feature vector before the LSH step. The statistics needed to compute the PCA weights are obtained by using feature vectors of images from the training data. Now, since each component of the feature vector is independent and has an equal chance of being positive or negative after the binarization step, two randomly chosen images have the least possible probability of having the same hash, specifically $2^{-k}$. To compare the similarity and distance between two hashes, we look at the Hamming distance ($L_1$ distance) between them. We quantify the degree of match between two hashes using the number of bits matched denoted by $M$ where

\vspace{-0.4cm}
\[
M(H(x),H(x')) = k - L_1(H(x), H(x'))
\]
\vspace{-0.6cm}
% The susceptibility of deep learning to adversarial attacks warrants a defense mechanism in our framework.

An adversarial attack involves adding a imperceptible amount of crafted perturbation of magnitude $\epsilon$ (typically $<8/255$) to each pixel of an image before passing it into the model \cite{subtle_adversarial}, which causes it to behave against its objective. This perturbation is generally created using information obtained from the model's gradients (white box attacks), which requires complete access to the model's architecture and weights. Other forms of attacks either require access to the model's dataset or output logits (gray box attacks) \cite{papernot2016distillationdefenseadversarialperturbations, inversion_attack, invert}, or only query access (black box attacks). There are two kinds of attacks relevant to perceptual hashes, hash collisions and hash aversions. A hash aversion attack would correspond to a malicious individual perturbing the original image $x$ to generate an image $x'$ that minimizes the similarity score $M$, leading to a false negative detection. A hash collision attack can by synthesized when an adversary has access to a a set of hashes stored by the CPO. This involves attacking a query image so that its hash matches one stored in the database, leading to a false positive detection. Adversarial training alone makes it extremely difficult for an adversary to perform either attack, even with full access to the model's weights. Furthermore, hash collision attacks are impossible to achieve in our framework due to the implementation of the secure MP-FHE scheme, which prevents the adversary's access to the perceptual hashes, even in the case of a data leak.

\citet{struppek2022learning, bhatia2022exploitingdefendingapproximatelinearity} have successfully demonstrated that both hash collisions and hash aversions can be performed against deep hashing networks, highlighting NeuralHash as a specific example. However, it is noteworthy that NeuralHash did not deploy any system to avoid these attacks. Adversarial training combined with the MP-FHE scheme secures our system against both attacks.

\subsection{Multi-Party Fully Homomorphic Encryption}
Multi-party Fully Homomorphic Encryption (MP-FHE) is a cryptographic technique that extends the capabilities of single-party homomorphic encryption to multiple participants. It enables computations being executed on encrypted data without revealing their individual inputs, while the decryption requires parties to perform collaboratively decrypt a result once the computation is complete. The advantage of this approach lies in its ability to preserve privacy while performing complex computations, such as those in secure MPC~\cite{Shamir}. Data confidentiality is an important goal in applications such as privacy-preserving data analysis~\cite{mouchet2021multiparty}.

MP-FHE also builds upon the concept of threshold cryptography, where the decryption process requires the collaboration of a threshold number of parties. This ensures that no single party can decrypt the data independently, enhancing security in scenarios involving untrusted parties. The MP-FHE scheme involves distributing the secret key across parties and utilizing homomorphic properties to perform computations on ciphertexts. Compared to traditional MPC methods like secret sharing, MP-FHE reduces communication overhead, as computations can be performed non-interactively on encrypted data, making it an efficient solution for large-scale, privacy-preserving applications~\cite{mouchet2021multiparty, lee2023efficient}.

\subsection{Deep Learning Based Detectors}

Synthetic images have artifacts that are distinctive to the generation process \cite{marra2019DoGAN, yu2019attributing}. Existing detectors often exploit these fingerprints in spatial \cite{sinitsa2023deep,liu2022detecting,cozzolino2018forensictransfer, marra2019incremental, du2020towards, jeon2020tgd} or frequency \cite{durall2020watch, frank2020leveraging, dzanic2020Fourier} domains to determine whether the image is synthetic or real. This method has two problems: a) Detectors trained on one type of generator do not generalize well on images from unseen generators, and b) transformations like resizing and compression destroy these fingerprints \cite{corvi2023detection}. Recent work exploits VLMs like CLIP \cite{radford2021learning} as feature extractors \cite{sha2023fake, amoroso2023parents, ojha2023towards,cozzolino2024raising} and train a classifier to detect synthetic images. \citet{cozzolino2024raising} clearly shows these models are more accurate and robust under transformations. We bechmark and analyse existing detectors in Appendix. We conclude leveraging VLM is the best choice for state of the art detection.

%% file: sec/4_proposed.tex
\section{Methodology}
The core database in the system is a content registry, indexed by content-producing organization~(CPO) with the perceptual hashes of content. The lifecycle of watermark creation consists of the content producing organization calculating perceptual hashes for all their content, encrypting those perceptual hashes to an MP-FHE encryption key for the database, and then digitally signing those encrypted perceptual hashes to establish provenance. Then, when users want to query the provenance of some potentially modified image, they request the parties running the MP-FHE database to find the images with the lowest Hamming distance, then are returned the most likely sources for the image. If no nearby matches are found, then the system falls back to the na\"ive AI image classifier model. 

\subsection{Perceptual Hash Design}

\citet{oquab2023dinov2} developed DINOv2, a self-supervised feature-extracting ViT network that is trained to generate feature vectors invariant to crops and natural perturbations like color jittering, gaussian blur and solarization.

We use the pre-trained DINOv2 ViT-B/14~(with register tokens) to extract feature embeddings of the given image. Then, we apply PCA to de-correlate each component of the feature vector and simultaneously reduce the dimensionality of the feature vector from $768$ to $96$, then apply the LSH step to obtain a $96$-bit binary string. Naturally, the resulting binary string serves the purpose of a perceptual hash since the original feature vector is highly robust to perceptual transformations of the image. To our knowledge we are also the first to develop an adversarial training scheme for deep perceptual hashing models, the details of which are discussed in \autoref{sec:adversarial}.

We also found that deep perceptual hashing models are, in general, more robust to attacks than classification models. This is due to the differing objectives of both adversaries, hence different image spaces available for attacks.

\noindent\textbf{Statistical Test.}
Let $H\in \{ 0,1 \}^{k}$ be the hash of an image $x$ in the database of the CPO that we wish to compare the query image against and, $H'$, be the corresponding hash of a query image $x'$. The detection test compares the number of matching bits between $H$ and $H'$ to a threshold, i.e. if
\begin{equation} 
M\left(H,H'\right) \geq \tau \,\,\textrm{ where }\,\, \tau\in 
\{0,\ldots,k\}, %\vspace{-0.3cm}
\end{equation}
then the two images are considered to be the same.

For the case where $x$ and $x'$ do not correspond to same image, we assume that bits $h_1,\ldots, h_k$ are i.i.d.  Bernoulli random variables with parameter $0.5$. Then $M(H, H')$ follows a binomial distribution with parameters~($k$, $0.5$).
% The binomial distribution can be approximated by the normal distribution $\mathcal{N}(0.5,\,\frac{0.25}{k})$ when $k$ is large. 

The False Positive Rate~(FPR) is the probability that $M(H, H')$ takes a value bigger than the threshold $\tau$.
\begin{align}
\label{eq:fpr}
    % \text{FPR}(\tau) & = \mathbb{P}\left(M \geq \tau \right) \approx \phi \left( \frac{0.5-\tau}{\sqrt{\frac{0.25}{k}}}\right). 
\text{FPR}(\tau) &= P(X \geq \tau), \quad X \sim \text{Binomial}(k, 0.5)
\end{align}

% Where $\phi$ denotes the cumulative density function of the standard normal variable.
% In the case of comparison against multiple images, we compare the hash $H'$ of the query image to the other hashes $ \left( H_{1},\dots, H_{N} \right)$. %
% There are now $N$ hypotheses to test, and if all are rejected, we conclude that the image was produced by the CPO.
% The FPR for this joint hypothesis for a given threshold $\tau$ is given by:
% \vspace{-0.2cm}
% \begin{equation}\label{eq:globalFPR}
%     \text{FPR}(\tau,N) = 1-(1-\text{FPR}(\tau))^N\approx N \cdot \text{FPR}(\tau).
% \vspace{-0.2cm}
% \end{equation}
Varying $\tau$ necessitates a trade-off between the False Negative Rate~(FNR) and the FPR, which can be set according to the sensitivity of the use case.

\subsection{Preserving Privacy of End-Users}
\noindent{\textbf{Problem Definition.}} In many real-world scenarios, maintaining a perceptual hash database poses significant privacy risks. Such a database could expose statistical information about end users and potentially would lead to multiple privacy breaches. For instance, leaking a perceptual hash of an AI generated image may allow partial prompt reconstruction, which can reveal personally identifiable information~(PII)~\cite{perceptual_hash_security_2024}. For example, widely used image generators like DALL-E are obliged to delete user data after up to 30 days, unless deidentified or retained for legal or security purposes~\cite{OpenAIHelp}. 
In another example, OpenAI's Advanced Voice mode is unavailable in the EU because the voice data is eligible for zero retention~\cite{OpenAIPolicies}. Thus, it is important to design a system for attesting the generated data by AI to preserve copyright related concerns and both preserve the content owners' privacy.
% Consequently, assuming a central and trustworthy database exists to compare new image queries is infeasible.

\noindent{\textbf{Design Principles.}}
To address this challenge, we need a system that can guarantee following:

\begin{compactitem}[-]
    \item \textbf{Data confidentiality:} No entity should gain direct access to any part of an image or information related to an image~(such as pHash) at any time.
    
    \item \textbf{Query Leakage resilience:} No query or database operation should lead to any data being revealed or stored~(temporary or permanently) on any device at any time.

    \item \textbf{Low Operational Risks:} Ideally, computations should require low security considerations and trustless scaling.
    
\end{compactitem}

% \noindent 
% Some of the guarantees outlined above can be achieved using a securely and privately shared database, as proposed in previous studies~\cite{edalatnejad2024janus, bloemen2024large}. However, certain principles, such as \textit{proof of match} and \textit{genuinity}, require our MPC setup to be publicly auditable~\cite{baum2014publicly}. To accomplish this, we propose using collaborative zkSNARKs~\cite{ozdemir2022experimenting}, enabling the verification of specific statements based on shared secrets without needing to reconstruct the shared secret. Additionally, to achieve \textit{private proofs of ownership}, we employ techniques suggested in prior work on proofs of provenance~\cite{vimz, veritas}.

\noindent\textbf{Adversary Model.}
We adopt a probabilistic polynomial-time~(PPT) adversary model, wherein any entity in the protocol may eavesdrop on or manipulate accessible data with polynomially bounded computational power.
Our constructions rely on several cryptographic primitives, which are assumed to be secure against any PPT adversary, specifically MP-FHE~\cite{mouchet2021multiparty}, hash functions, and digital signatures.

\begin{figure}[t]
  \centering
  \includegraphics[width=0.9\linewidth,trim={.5cm 0 0 0},clip]{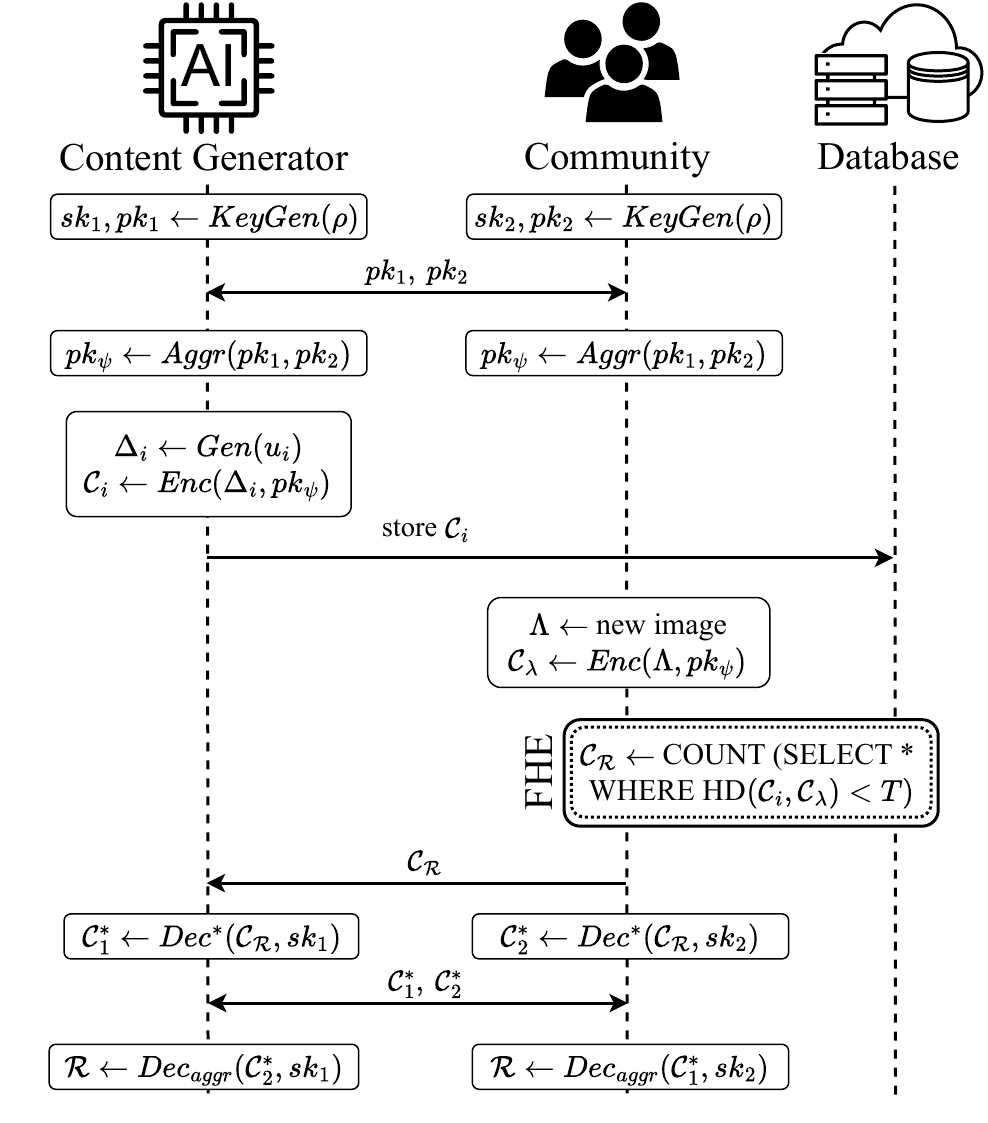}
   \caption{\textbf{Overview of the protocol.} The database is kept private, content generators can write encrypted data to it, and the public can run private queries that return only provenance data.}
   \label{fig:protocol}
   \vspace{-0.5cm}
\end{figure}

% In this section we provide a general overview of the proposed architecture. The solution illustrates how a Multi-Party Fully Homomorphic Encryption~(MP-FHE) scheme can be integrated into the Proteus protocol to ensure preserving privacy.

\vspace{-0.5cm}
\paragraph{\textbf{Protocol Details}}
\autoref{fig:protocol} illustrates a scenario involving two parties~(for simplicity) executing a secure protocol: one as the content creator~(the owner of generative model) and the other as a regulator or trusted entity~(a community representative). Each party possesses their own set of private and public keys. The protocol operates in the following phases:
\begin{compactitem}[-]
    \item \textbf{Setup Phase:} Parties exchange shares of their public keys to create an aggregated $m\mbox{-of-}n$ MPC public key. The shared key is used universally to encrypt pHash values. Note that all parties generating and signing data are incentivized to contribute up to $(n - m)$ live servers to the protocol, so that they can help ensure that no one decrypts database values and people conform to the protocol. More servers means they could affect liveness, and less servers means they have less control over decryption. 
    
    \item \textbf{Encryption Phase:} pHash value of each image is encrypted by the generator using the aggregated public key~($P_{shared}$) and stored in database.
    To enhance the accuracy of future claims of copyright violations, parties may decide to also store MPC-encrypted version of each image for a ceremony-based opening under specific conditions, like when pHash matches show very low hamming distance. 
    
    \item \textbf{Evaluation Phase:} When a new image is submitted for evaluation, the pHash value of the image is encrypted using the same shared key~($P_{shared}$) and undergoes a fully homomorphic computation of the Hamming distance across the database. The result is an encrypted output indicating the number of pHash values with close Hamming distances.
    The precise FHE function may vary depending on the specific scenario, database size, or the quality of the pHash values. 
    
    \item \textbf{Decryption Phase:} To decrypt evaluation results, both parties exchange decryption shares in an MPC process. The final value reveals whether the image has any overlaps in the database~(non-zero result) or not~(zero result).
\end{compactitem}

\noindent We implemented our proof-of-concept (PoC) for the proposed system using the \emph{Phantom-Zone} MP-FHE library\footnote{\url{https://github.com/gausslabs/phantom-zone}} in FHEW and link to a comparable implementation of dot product in BFV in Lattigo\footnote{\url{https://eprint.iacr.org/2024/1774.pdf}}.

One of the key security guarantees of the protocol is that the queries do not leak any information about the dataset, beyond confirming whether the pHash of the query image shares significant similarities with any value in the dataset. To achieve this, we define a set of valid queries that are allowed, with the database managed in an MPC manner.

\subsection{Detector for Unknown Images}

\noindent\textbf{Architecture} Open Clip~\cite{ilharco_openclip_2021} has trained various contrastive image-language models with better zero-shot performance on ImageNet \cite{imagenet} than OpenAI's open-sourced version. So, we choose four image encoders for our experiments: SigLIP \cite{zhai2023sigmoid} models with SoViT-400M \cite{alabdulmohsin2024getting} backbone with an input resolution of 224 and 384 and CLIP models with ViT-H \cite{dosovitskiy2020vit} backbone with an input resolution of 224 and 378. We use SVM as our classifier and the second last layer to extract embeddings following \cite{cozzolino2024raising}. We did an extensive ablation study to decide the exact architecture of our model. Results of the study are in Appendix.

\noindent\textbf{Method} We perform two sets of experiments. 1) We use the better encoders from Open CLIP and train an SVM on the features extracted from their visual encoders. 2) We finetune the entire encoder backbone along with the SVM layer. 
% We follow the technique proposed by \citet{wortsman2022robust} and interpolate between the finetuned and original encoder weights for robustness.
In the first set of experiments we train the SVM on frozen encoder features for 20 epochs with a batch size of 32. In the second set of experiments. We find that the training loss converges significantly within the first two epochs. So, we finetune each model for five epochs and use the validation set to choose the epoch, after which we get the best-performing model. \Cref{tab:training_config} details the finetuning configurations for each model. We also perform adversarial training to protect from malicious attack. Details of adversarial training and accuracy scores are in Appendix.

%% file: sec/5_privacy_preserving.tex
\begin{figure}[!t]
    \vspace{-0.5cm}
    \centering
    \clearpage
    \includegraphics[page=1, width=\linewidth]{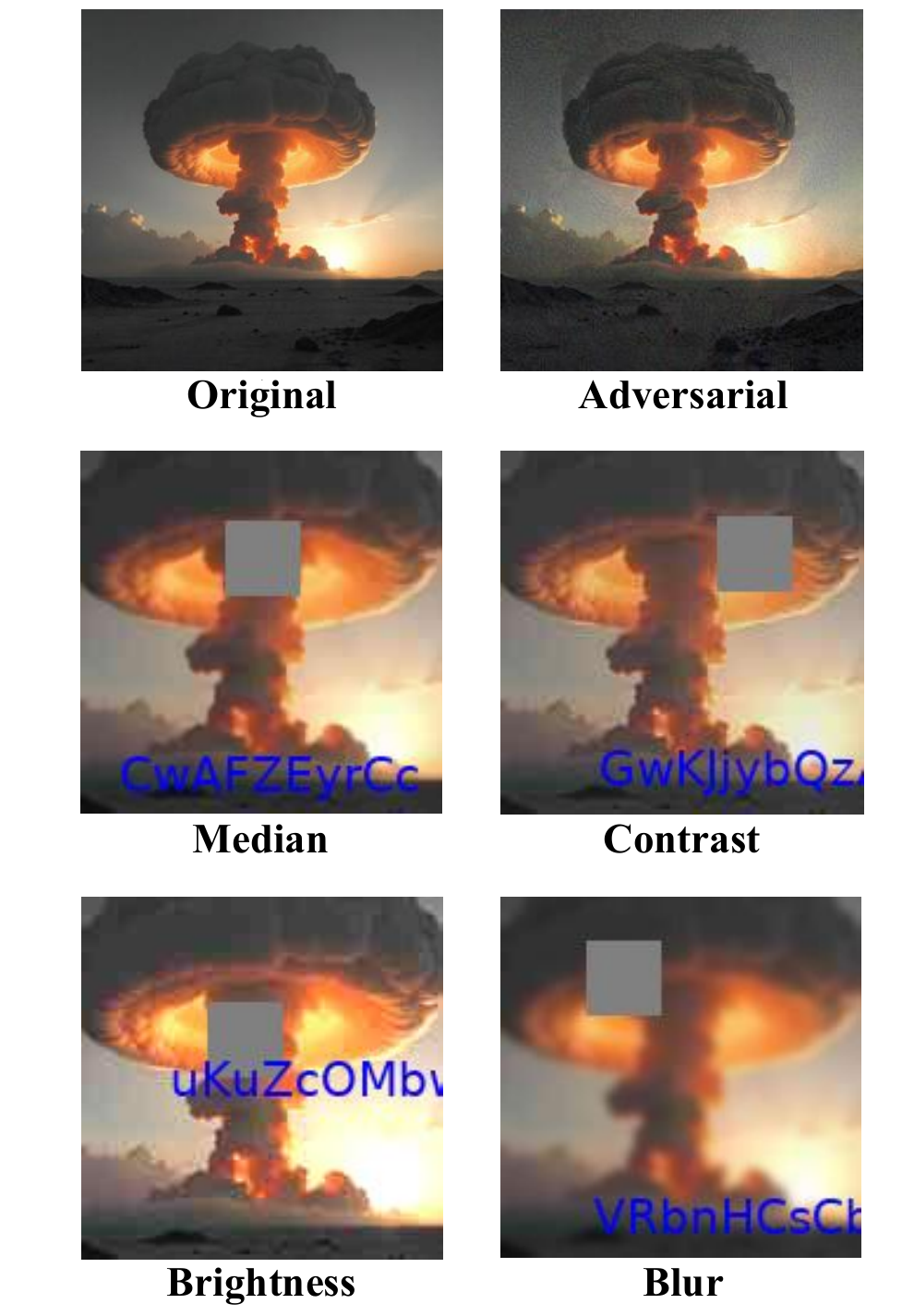}
    \vspace{-0.5cm}
    \caption{Transformations used to benchmark hashing and watermarking methods. With screenshot simulation as pre-processing, and random erasing followed by text overlay as post-processing.}
    \label{fig:examples}
    \vspace{-0.5cm}
\end{figure}

\section{Experimental Results}
In this section, first we evaluate the robustness and accuracy of our proposed DinoHash and DL detector. 
Finally, we evaluate our PoC of the protocol using the Phantom Zone FHEW library~\cite{PhantomZone} across different setups, ranging from midrange laptops to servers in \autoref{sec:benchmark-fhew}.

\subsection{DinoHash}
We benchmark DinoHash against other prominent methods like NeuralHash, Stable Signature, and DCT-DWT, across multiple transformations designed to mimic typical image edits. We run all experiments on a single NVIDIA GPU A100 GPU with 40GB of VRAM. We present our results for four different complex transformations, each transformation consists of a base transformation of a $20\%$ crop from each side followed by a JPEG compression with a quality factor of $30\%$. This simulates an effect similar to a screenshot. We then apply one of the following four operations: 

\begin{compactenum}
    \item a brightness shift by a random factor between $\pm 30\%$
    \item a contrast shift by a random factor between $\pm 30\%$
    \item gaussian blurring with a kernel width of 2
    \item application of a median filter with $k=3$
\end{compactenum}
\noindent These represent common image transformations and filters found in most image editing applications. To ensure even stronger robustness against complex image distortions: we then apply a random-erasing augmentation, where a square area with a side length $20\%$ of the image is erased; followed by a random text overlay with a random string of $10$ alphanumeric characters. A visualization of these transformations can be found in \autoref{fig:examples}.

\begin{figure}[h] 
    \centering

\begin{tikzpicture} 
\hspace{-0.5cm}
\begin{axis}[
    width=0.48\textwidth,
    height=0.2\textwidth,
    ylabel={Bit Accuracy (\%)},
    xtick=data,
    symbolic x coords={Brightness, Contrast, Median, Blur},
    ybar=0.2,
    bar width=0.25cm,
    enlarge x limits={abs=0.6cm},
    legend style={
        at={(0.5,-0.45)},
        anchor=north,
        legend columns=2,
        column sep=0.5cm
    },
    ymin=48,
    ymax=85,
    grid=both
]
    % dct-dwt
    \addplot[fill=red!40] coordinates {
        (Brightness, 50)
        (Contrast, 50) 
        (Median, 50)
        (Blur, 50)
    };
    
    % Stable Signature
    \addplot[fill=blue!40] coordinates {
        (Brightness, 78.11)
        (Contrast, 78.14) 
        (Median, 71.7)
        (Blur, 49.38)
    };

    % NeuralHash
    \addplot[fill=orange!40] coordinates {
        (Brightness, 74.51)
        (Contrast, 74.79) 
        (Median, 74.96)
        (Blur, 74.35)
    };
    
    % DinoHash
    \addplot[fill=green!40] coordinates {
        (Brightness, 82.64)
        (Contrast, 82.69) 
        (Median, 82.45)
        (Blur, 79.63)
    };
    
    \legend{DCT-DWT, Stable Signature, NeuralHash, DinoHash}
\end{axis}
\end{tikzpicture}

\caption{\textbf{Bit Accuracy Comparison.} The mean bit accuracy of different methods after a simulated screenshot followed by the respective transformations.}
\label{fig:barplot}
\vspace{-0.5cm}
\end{figure}

\subsection{Distortion Robustness}

We use 1M randomly sampled images from the DiffusionDB dataset \cite{wang2023diffusiondblargescalepromptgallery} to carry out experiments for DCT-DWT watermarking, NeuralHash and our framework. We use 1M randomly sampled prompts of the dataset to generate images to benchmark Stable Signature.

\begin{figure}[!h] 
    \centering
    \includegraphics[width=0.5\textwidth]{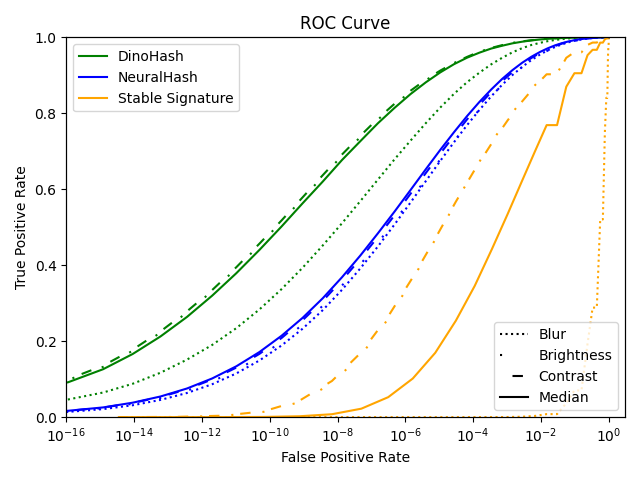}
    \caption{\textbf{Detection Results.} The TPR/FPR detection curve of different methods after a simulated screenshot followed by the respective transformations, random erasing and text overlay.}
    \label{fig:roc}
    \vspace{-0.5cm}
\end{figure}

\autoref{fig:roc} shows the Receiver Operating Characteristic (ROC) Curve for different hashing/watermarking methods under different transformations. The level of $\tau$ is varied from $0$ to $96$ to obtain different values for the TPR and FPR. The TPR is calculated by measuring the fraction of images whose degree of match, $M$, after the transformation equals or exceeds $\tau$. Conversely, the FPR is approximated using \autoref{eq:fpr} since it is too small to be measured experimentally. \autoref{fig:barplot} reports results on the mean bit accuracy of the algorithms. We do not include the TPR-FPR plot for DCT-DWT due to its near-random performance.

We observe that all algorithms are roughly equally robust to the brightness and contrast transformations due to their non-destructive nature. We also observe that DinoHash is significantly more robust to transformations than NeuralHash and Stable Signature across both metrics. We do not report the results for the vanilla (no transformation) case since any hashing method would trivially achieve a $100\%$ match in the cases without perturbations. This, however, does not always hold for watermarking methods since the watermark generation and extraction models may not always agree. Furthermore, all watermarking methods report the PSNR (Peak Signal-to-Noise Ratio) and SSIM (Structural Similarity Index) of the watermarked images. These metrics measure how perceptually similar the watermarked and the unwatermarked images are. Since hashing algorithms do not cause any modification to the underlying image, we achieve perfect PSNR and SSIM scores. Since NeuralHash is a proprietary algorithm, it is difficult to comment on why DinoHash performs better.

\subsection{Experiments on DL detector}
\label{subsec:exp_dl}
\begin{table*}[t]
\centering
% \small
\begin{adjustbox}{width=0.95\linewidth}  
\setlength{\tabcolsep}{6pt} % Adjust column padding as needed
\begin{tabular}{lcccccccc}
\toprule
{Method} & {Backbone} & {Resolution} & DALL-E 2 & DALL-E 3 & Firefly & Midjourney v5   & {AVG} \\ 

\midrule
\citet{cozzolino2024raising} & - & - & 0.706 / \textbf{0.612} & 0.794 / 0.639 & 0.766 / 0.635 & 0.698 / 0.602 & 0.741 / 0.622 \\
\citet{corvi2023detection} & - & - &  0.689 / 0.502 & 0.287 / 0.500 & 0.574 / 0.501 & 0.481 / 0.500 & 0.508 / 0.501 \\
\midrule
\textit{Ours}-frozen \\
\midrule
DFN \cite{fang2023data} & ViT-H & 378 & 0.767 / 0.584 & \textbf{0.999} / \textbf{0.980} & \textbf{0.936} / \textbf{0.754} & \textbf{0.949} / \textbf{0.804} & \textbf{0.913} / \textbf{0.780} \\
DFN \cite{fang2023data} & ViT-H & 224& \textit{0.792} / \textit{0.605} & 0.997 / \textit{0.974} & 0.921 / 0.720 & 0.934 / 0.771 & \textit{0.911} / 0.767 \\
SigLIP\cite{zhai2023sigmoid} & So-ViT-400M & 224 & 0.663 / 0.574 & 0.996 / 0.965 & 0.896 / 0.730 & 0.909 / 0.785 & 0.866 / 0.764 \\
SigLIP\cite{zhai2023sigmoid} & So-ViT-400M & 384 & 0.731 / 0.600 & 0.997 / 0.972 & \textit{0.924} / \textit{0.742} & 0.926 / 0.772 & 0.895 / \textit{0.771} \\
\midrule
\textit{Ours}-finetuned \\
\midrule
DFN \cite{fang2023data} & ViT-H & 378 & 0.742 / 0.534 & 0.991 / 0.965 & 0.870 / 0.585 & 0.906 / 0.706 & 0.877 / 0.697 \\
DFN \cite{fang2023data} & ViT-H & 224 & \textbf{0.818} / 0.559 & \textit{0.998} / 0.972 & 0.910 / 0.623 & 0.880 / 0.689 & 0.901 / 0.711 \\
SigLIP\cite{zhai2023sigmoid} & So-ViT-400M & 224 & 0.781 / 0.542 & 0.997 / 0.966 & 0.887 / 0.620 & 0.937 / 0.737 & 0.901 / 0.716 \\
SigLIP\cite{zhai2023sigmoid} & So-ViT-400M & 384 & 0.772 / 0.592 & 0.993 / 0.966 & 0.903 / 0.664 & \textit{0.938} / 0.777 & 0.902 / 0.750 \\
\bottomrule
\end{tabular}
\end{adjustbox}
\caption{AUC/Accuracy scores for various methods across different commercial tools with average values. Accuracy values here are calculated with threshold of 0.5. \textbf{Bold} represent highest value in a column and \textit{italics} represent second highest value.}
\label{tab:auc_acc_table}
\end{table*}
\vspace{-5pt}
% \begin{table*}[t]
% \centering
% % \small
% \begin{adjustbox}{width=0.95\linewidth} 
% \setlength{\tabcolsep}{5pt} % Adjust column padding as needed
% \begin{tabular}{lccccccccc}
% \toprule
% {Method} & {Backbone} & {Resolution} &{Patch Size} & DALL-E 2 & DALL-E 3 & Firefly & Midjourney v5 & {AVG} \\ 

% \midrule
% \citet{cozzolino2024raising} & - & - & -& \textbf{0.612} & 0.639 & 0.635 & 0.602 & 0.622 \\
% \citet{corvi2023detection} & - & - &-& 0.502 & 0.500 & 0.501 & 0.500 & 0.501 \\
% \midrule
% \textit{Ours}-frozen \\
% \midrule
% DFN \cite{fang2023data} & ViT-H & 378 &14& 0.584 & \textbf{0.980} & \textbf{0.754} & \textbf{0.804} & \textbf{0.780} \\
% DFN \cite{fang2023data} & ViT-H & 224 &14& \textit{0.605} & \textit{0.974} & 0.720 & 0.771 & 0.767 \\
% SigLIP\cite{zhai2023sigmoid} & So-ViT-400M & 224 &14& 0.574 & 0.965 & 0.730 & \textit{0.785} & 0.764 \\
% SigLIP\cite{zhai2023sigmoid} & So-ViT-400M & 384 &14& 0.600 & 0.972 & \textit{0.742} & 0.772 & \textit{0.771} \\
% \midrule
% \textit{Ours}-finetuned \\
% \midrule
% DFN \cite{fang2023data} & ViT-H & 378 &14& 0.534 & 0.965 & 0.585 & 0.706 & 0.697 \\
% DFN \cite{fang2023data} & ViT-H & 224 &14& 0.559 & 0.972 & 0.623 & 0.689 & 0.711 \\
% SigLIP\cite{zhai2023sigmoid} & So-ViT-400M & 224&14 & 0.542 & 0.966 & 0.620 & 0.737 & 0.716 \\
% SigLIP\cite{zhai2023sigmoid} & So-ViT-400M & 384 &14& 0.592 & 0.966 & 0.664 & 0.777 & 0.750 \\
% \bottomrule
% \end{tabular}
% \end{adjustbox}
% \caption{Accuracy values with 0.5 as threshold for various methods across different models with average values.}
% \label{tab:acc_table}
% \end{table*}

We fix a consistent setup to simulate real world conditions on a challenging dataset with postprocessing steps to evaluate our models as accurately as possible.

\noindent\textbf{Dataset} We report all results on the recently proposed Synthbuster dataset \cite{bammey2023synthbuster} as it has images from widely used commercial tools, i.e. Dalle2 \cite{dalle2}, Dalle3 \cite{dalle3}, Midjourneyv5 \cite{midjourney} and Adobe Firefly \cite{firefly}. It uses RAISE-1k \cite{dang2015raise} as its real part. Each source contains 1000 images. We use 10000 pairs of real and synthetic images for all training purposes. This set is created using the strategy proposed in \cite{cozzolino2024raising}. We randomly select 10000 images from the train set of COCO-2017\cite{lin2014microsoft}, generate their caption using BLIP \cite{li2022blip}, and generate images for these captions using Stable Diffusion XL \cite{podell2023sdxl}. We use a validation set of 500 pairs of images created similarly using the validation set of COCO-2017.
 
\noindent\textbf{Transformations} In real life, images often suffer cropping and compression when shared. We perform a set of transformations on all images to make the model more robust and to test its robustness under such attacks. We perform random crop to 62.5\% to 100\% of the original scale followed by resizing to 200x200 pixels. We then perform random jpeg compression to between 85\% to 100\% of initial quality.

\noindent\textbf{Testing Protocol} Our testing dataset contains synthetic images from four different sources having 1000 images each and a real set of 1000 images. While computing metrics for one synthetic image source, we consider just that subset of dataset and the real images. So, we effectively have 1000 synthetic and real images, making the dataset balanced.

\noindent\textbf{Metrics} We use area under ROC curve (AUC) and accuracy (ACC) with threshold of 0.5 as our metric. We use threshold of 0.5 to simulate a situation with no prior information about the detector.

\noindent\textbf{Results} Our models outperform the previous SoTA~\cite{cozzolino2024raising} across both metrics. Linear SVM trained on top of frozen CLIP image encoder with ViT-H backbone, 378 input resolution and quickGELU activation function yields best results on most generators and on average. It shows improvements of 23.3\% / 25.4\% in AUC/ACC respectively over previous SoTA. Other models also perform better than~\citet{cozzolino2024raising}, we present detailed results in~\cref{tab:auc_acc_table}.
% and~\cref{tab:acc_table}.

% We also tried zero shot experiments leveraging the text encoder of CLIP for classification, but results were poor. More details in appendix. 

\begin{table}
\centering
\footnotesize
\setlength{\tabcolsep}{5pt} % Adjust column padding as needed
\begin{tabular}{lccccc}
\toprule
{Method} & {Backbone} & {Resolution} & {Epoch} & {LR} &{Batch Size} \\
\midrule
SigLIP \cite{zhai2023sigmoid} & So-ViT-400M & 384 & 5 & 5e-6& 8 \\ 
SigLIP \cite{zhai2023sigmoid} & So-ViT-400M & 224 & 3 & 5e-6&8 \\
DFN \cite{fang2023data} & ViT-H-14 & 378 & 4 & 5e-6&8 \\ 
DFN \cite{fang2023data} & ViT-H & 224 & 3 & 5e-6&8 \\ 
\bottomrule
\end{tabular}

\caption{Training configurations for various methods, including backbone, resolution, epochs, and learning rate.}
\label{tab:training_config}
\end{table}
\vspace{-10pt}

%% file: sec/6_conclusion.tex
\vspace{0.2cm}
\section{Limitations and Future Work}
\label{sec:future}
Incorporating FHE into the querying process increases computational overhead, which could impact performance in real-time applications. Future research could focus on optimizing FHE-based indexing methods for quick retrieval to meet computational demands, or augmenting the Worldcoin Iris matching system or Apple's Wally for our use case.

One advantage of watermarking over hashing is that watermarking only requires storing \textit{one hash per image creator} for provenance, as it needs only to identify the image’s creator, not each specific image. In contrast, perceptual hashing requires storing \textit{one hash per image}, which increases the search space to make matching more robust but requires more memory and increased latency.

\vspace{-10pt}
\vspace{0.2cm}
\section{Conclusion}
Our three-part framework addresses significant challenges in AI content provenance detection by combining perceptual hashing, FHE, and a robust AI content detection model. Our perceptual hashing method with DINOV2 overcomes the limitations of traditional watermarking by capturing semantic and structural image features that persist through common transformations without altering the base image while also maintaining robustness to adversarial attacks. Applying FHE within this framework ensures that users can query content databases securely, preserving privacy without compromising the system's integrity, even if some data is exposed. Finally, the AI-generated content detection model increases the framework’s robustness by identifying AI-generated images that may be generated outside the known database.

\clearpage

%% file: sec/appendix.tex
\clearpage
\setcounter{page}{1}
\maketitlesupplementary
\def\phiorg{\mathop{\phi_{\rm{Org}}}\nolimits}

\section{Benchmarking existing DL detectors}
\label{sec:benchmarking}
\begin{figure}[!t]
    \centering
    \includegraphics[width=0.95\linewidth]{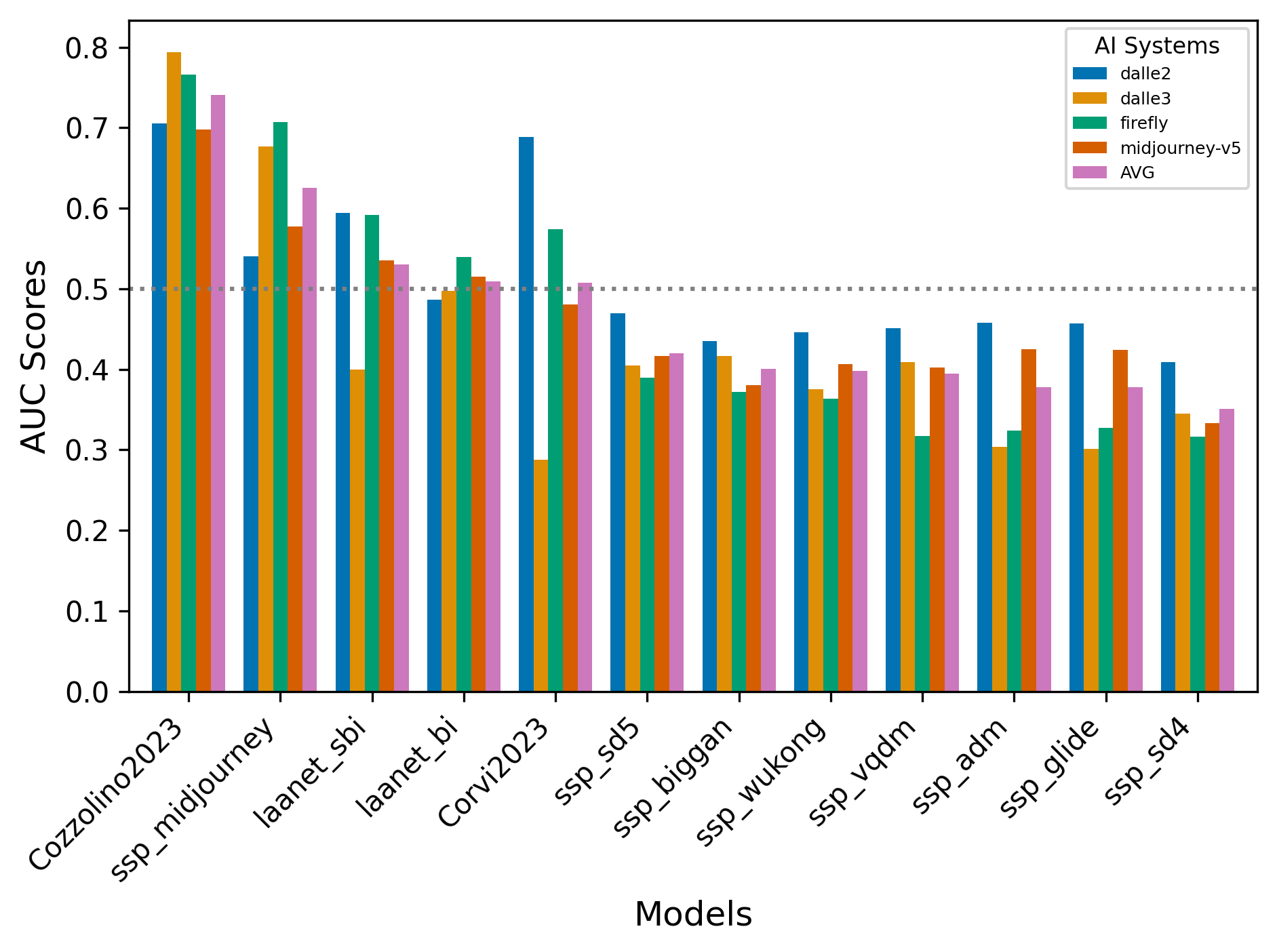}
    \caption{Area Under ROC curve for various models on postprocessed images. Arranged in descending order of average AUC.}
    \label{fig:auc_bench}
\end{figure}

Existing litreature contains various methods for detecting AI generated images using machine learning models. These methods can be classified primarily into a) Artifact based methods and b) VLM based methods. Artifact based methods \citep{wang2020cnn,gragnaniello2021gan,mandelli2022detecting,chai2020makes,nguyen2024laa} leverage fingerprints of image generators or noise patterns of camera \citep{chen2024single} to detect AI generated image. While this method is effective, it often does not generalize well to unseen generators and artifacts are often lost under operations like JPEG compression, resizing and random crop. VLM based methods \citep{ojha2023towards,cozzolino2024raising} leverage the join image-language feature space to detect fake images.

\begin{figure}[!t]
    \centering
    \includegraphics[width=0.95\linewidth]{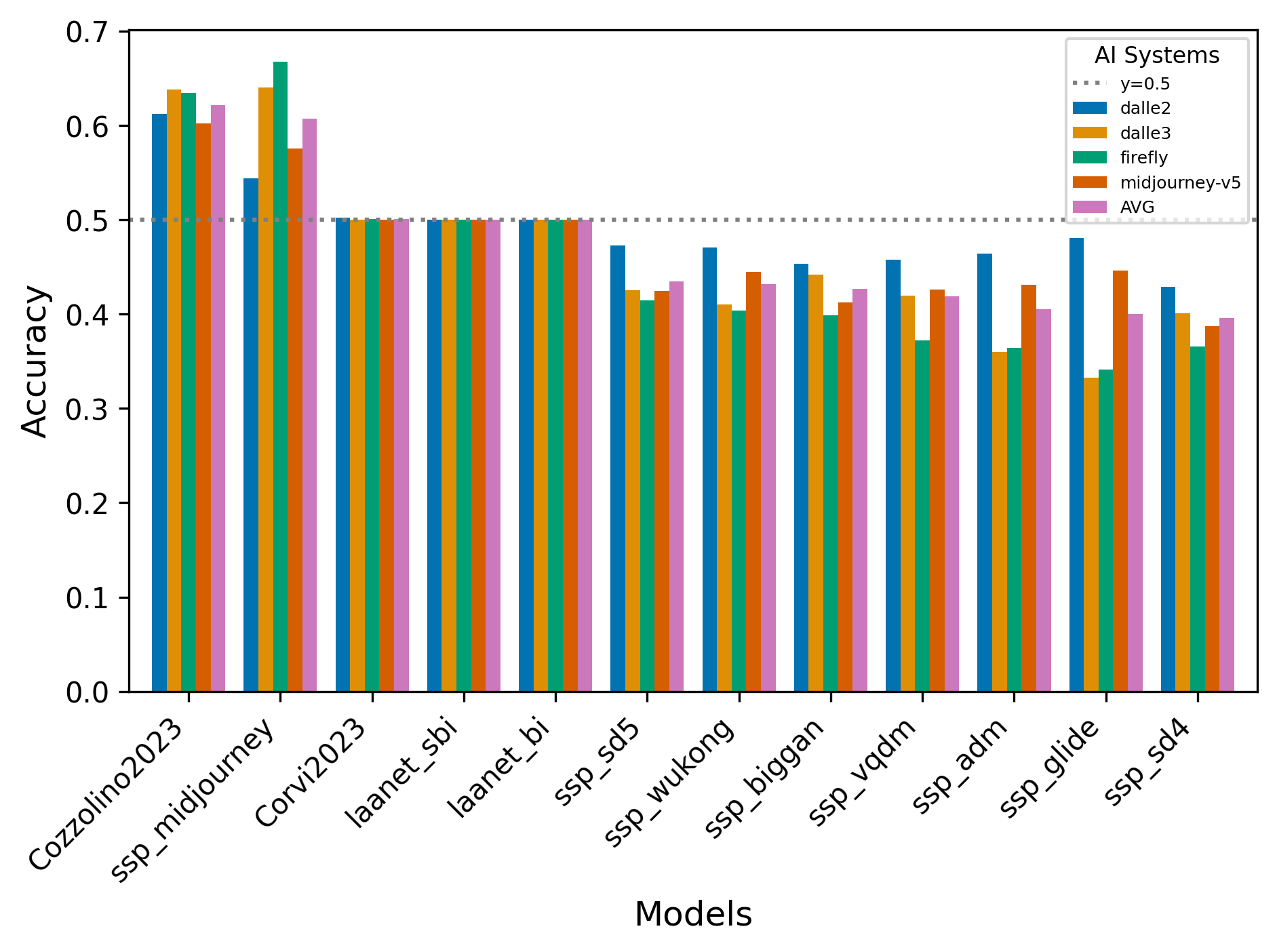}
    \caption{Accuracy for various models on postprocessed images. Arranged in descending order of average accuracy.}
    \label{fig:acc_bench}
\end{figure}

In this era of social media and freely available text-to-image generators the task of detection AI generated images has gained a lot of attention. A lot of different approaches are available but we found a lack of common benchmark. A lot of authors often do not report results on realistic scenarios like highly compressed images. \citet{cozzolino2024raising} benchmark state-of-the-art detectors and show through extensive experiments that VLM based method perform well on postprocessed images while performance of artifact based methods often fall to near random chance. Despite this effort there still existed new detectors \citep{nguyen2024laa,chen2024single} that were not benchmarked under the same condition. We benchmark all detectors on the dataset and under transformations described in \Cref{subsec:exp_dl}. The Area Under Curve and Accuracy for the models not benchmarked in \citep{cozzolino2024raising} are visualized in \Cref{fig:auc_bench} and \Cref{fig:acc_bench} respectively. Our benchmarking reinforced the findings of \citet{cozzolino2024raising} and shows the superiority of VLM based detectors. Hence, we choose to leverage VLM based detector in our framework.

\section{Ablation Study}
\label{sec:ablation}
\begin{table*}[h]
    \centering
    \small
    \begin{tabular}{lccccccccc}
        \toprule
        VLM & Backbone&Resolution & Classifier & Text & DALL·E 2 & DALL·E 3 & Firefly & Midjourney v5 & Avg \\
        \midrule 
        CLIP & ViT-L & 224& Zero-shot & Fixed & 0.524&0.558&0.550&0.534&0.542 \\
        CLIP & ViT-L & 224& Zero-shot & BLIP & 0.508&0.476&0.475&0.474& 0.483 \\
        DFN & ViT-H & 224& Zero-shot & BLIP &0.609&0.818&0.670&0.663 & 0.690 \\
        CLIP & ViT-L & 224 & SVM & - &  0.680&0.74&0.646&0.627&0.673 \\
        DFN & ViT-H & 224& SVM & - & \textbf{0.792} & \textbf{0.997} & \textbf{0.921} & \textbf{0.934} & \textbf{0.911} \\
        SigLIP & so-ViT-400M & 224& SVM & - & 0.663 & 0.996 & 0.896 & 0.909 & 0.866 \\
        \bottomrule
    \end{tabular}
    \caption{Ablation study on AI detection model architecture with individual model performance. Table contains accuracy values on non postprocessed images.}
    \label{tab:ablation_expanded}
\end{table*}

\Cref{sec:benchmarking} clearly motivates the use of VLMs for generalizable and robust detector. Choice of architecture and exact method of detection can significantly affect detection. We have to choose which VLM to use, size of backbone (B/L/H) to use and the method of detection.  We can either use zero shot capablities of VLMs or train a classifier on top of features extracted from VLMs. If we choose zero shot methods we could either use fixed text template or generate custom text using state of the art models like BLIP.
We explain each choice in detail below.

\subsection{Choice of Backbone}
\citet{radford2021learning} introduced contrastive learning of joint language image embedding space. The pioneering CLIP model was adopted and used extensively due to its generalization and few shot capabilities. Follow-up work \citep{fang2023data,zhai2023sigmoid} improved this through various strategies, including additional loss terms, a new dataset created using novel data filtering strategies and changes in normalization. Open CLIP \citep{ilharco_openclip_2021} has trained a range of these models on varying backbones, datasets and resolutions. We find that DFN \citep{fang2023data} and SigLIP-based models \citep{zhai2023sigmoid} have the highest zero-shot imagenet accuracies. Our task involves using these models for downstream tasks, so we use zero-shot ImageNet accuracies to indicate which model could perform well on our task. Thus, we experiment with DFN and SigLIP-based models along with Open AI's CLIP. We have three different backbones, i.e. ViT-L, ViT-H and so-ViT-400M. Results are detailes in \Cref{tab:ablation_expanded}. We observe that the DFN-based ViT-H backbone outperforms the CLIP-based ViT-L backbone and SigLIP-based so-ViT-400M backbone when using the same classification strategy, i.e. SVM. It also outperforms CLIP-based ViT-L when we do zero-shot classification. So, the DFN-based ViT-H backbone is the clear choice for our detection model. We note that the SigLIP-based so-ViT-400M backbone has performance comparable performance, so we include it in further experiments.

\subsection{Method of Detection}
\citet{cozzolino2024raising} tested Support Vector Machine, Logistic Regression (LR), Mahalanobis distance (MAH), Gaussian Naive Bayes classifier
(GNB), Soft voting k-Nearest Neighbor (SNN) with Open AI's CLIP and found out that SVM performs the best on the task of detection AI generated images. However, CLIP's zero shot capablities were not tested. So we test three classification strategies, namely zero-shot classification with fixed text template, zero-shot classification with BLIP generated text template and SVM. \Cref{tab:ablation_expanded} clearly shows SVM performs better, so we use SVM in our final detector.

\section{Adversarial Training}\label{sec:adversarial}

In this section we discuss the adversarial fine-tuning implementation details for DinoHash and DL based detector.

\subsection{DinoHash}
Given a hashing model, $H$, we would like to finetune it such that an adversarial perturbed image, $x'$, produces the same hash as the original image, $x$ where $\norm{x-x'}_\infty \leq \epsilon$, i.e. we would like that $H(x') \approx H(x)$. Hence, we initially formulated the following objective for finetuning:

\begin{align}
H_{robust}=&\argmin_{H}\sum_{i=1}^n \max_{\norm{x_i'-x_i}_\infty \leq \epsilon} \norm{H(x_i')-H(x_i)}_1.
\end{align}

The downside of this approach was that it led to a collapse of the hashing function and caused it to produce the same hash for all images, which is undesirable. This behaviour can be explained since the proposed loss function is perfectly minimized. This led us to formulate our second optimization problem:

\begin{align}
H_{robust}=&\argmin_{H}\sum_{i=1}^n \max_{\norm{x_i'-x_i}_\infty \leq \epsilon} \norm{H(x_i')-H_{orig}(x_i)}_1.
\end{align}

Where $H_{orig}$ denotes the original, non-robust hashing model. This ensured that the model retains its original hashing capability. The inner maximization problem was solved by adapting APGD \citenum{autoattack}, a popular algorithm for adversarial attacks, to our use case.

Since the heaviside step function is non-differentiable, we used cross-entropy loss on the logits before the binarization step. Furthermore, we observed that adding a clean-loss term aided the optimization process and resulted in faster convergence. The final loss function used in code was:

\begin{equation}
\begin{split}
L(x_i)= \text{CE}(\hat{H}(A(x_i)), \: \hat{H}_{orig}(x_i)) \: + \\
\alpha \cdot \text{CE}(\hat{H}(x_i), \: \hat{H}_{orig}(x_i))
\end{split}
\end{equation}

Here $A(x)$ represents the attacked version of image $x$ generated by APGD with $\epsilon=8/255$, CE represents the cross-entropy loss with the second term as the target logit, $\hat{H}$ represents the deep hashing model without the binarization step, and $\alpha$ represents the loss weightage for clean examples and was set to $500$ in our experiments. 
\begin{table*}[htbp]
    \centering
    \begin{tabular}{l c c c c c c}
        \toprule
       ` & \multicolumn{2}{c}{Clean Data} & \multicolumn{2}{c}{Adversarial} & \multicolumn{2}{c}{Adversarial} \\
        \cmidrule(lr){2-3} \cmidrule(lr){4-5} \cmidrule(lr){6-7}
        $\boldsymbol{\varepsilon}$
 & \multicolumn{2}{c}{-} & \multicolumn{2}{c}{4} & \multicolumn{2}{c}{8} \\
        \midrule
        & Non-Robust & Robust & Non-Robust & Robust & Non-Robust & Robust \\
        \midrule
        DALL·E 2 & 0.559 & 0.54 & 0.645 & 0.838 & 0.662 & 0.822 \\
        DALL·E 3 & 0.972 & 0.847 & 0.66 & 0.867 & 0.685 & 0.864 \\
        Firefly & 0.623 & 0.574 & 0.632 & 0.826 & 0.641 & 0.825 \\
        Midjourney v5 & 0.689 & 0.614 & 0.651 & 0.85 & 0.662 & 0.838 \\
        Average & 0.711 & 0.644 & 0.647 & 0.845 & 0.663 & 0.837 \\
        \bottomrule
    \end{tabular}
    \caption{Comparison of Non-Robust and Robust Models Across Attack Types and Generators}
    \label{tab:robust_dldetector}
\end{table*}
We fine-tune DinoHash on DiffusionDB using the loss function defined above for $20,000$ steps with a batch size of $256$ using the AdamW optimizer. The adversarial examples were generated using APGD with $10$ steps.

We compare the adversarial robustness of DinoHash against different baseline models in \autoref{tab:adv_hash}. We note that for all baseline models, the adversarial attack achieves near-perfect accuracy, significantly underscoring the insecurity of their application as compared to DinoHash. 

We would like to mention that we were initially experimenting with DISCO \cite{disco} to enhance robustness against adversarial attacks. However, our experiments indicated that DISCO is extremely vulnerable to standard PGD attacks, despite prior claims that the Backward Pass Differentiable Approximation Attack \citenum{obfuscated} was the strongest. This suggests that its reported robustness is heavily overstated.

\renewcommand{\arraystretch}{1.5}
\begin{table}[h]
    \centering
    \begin{tabular}{cccc}
        \toprule
        $\epsilon$ & DinoHash & NeuralHash & Stable Signature \\
        \midrule
        $\frac{4}{255}$ & $\textbf{62.0\%}$ & $0.1\%$ & $0.0\%$ \\
        $\frac{8}{255}$ & $\textbf{64.8\%}$ & $0.0\%$ & $0.0\%$ \\
        \bottomrule
    \end{tabular}
    \caption{Performance of different models under adversarial perturbations with the APGD attack performed using $100$ steps}
    \label{tab:adv_hash}
\end{table}

We also found that deep perceptual hashing models are, in general, more robust to attacks than classification models. This is due to the differing objectives of both adversaries, hence different image spaces available for attacks. Take a trained classifier with $1000$ output logits and a hashing model with a $1000$-dimensional hash as well. To successfully perform an attack on the classifier, the adversary needs to find an input image that perturbs only one logit, the one corresponding to the correct class. On the other hand to successfully attack a hashing model the adversary needs to perturb \textbf{atleast} $500$ logits, leading to a much tighter search-space. To validate this claim we varied the resulting dimension of the PCA post-processing step and observed that as the number of dimensions of the hash increased, the attack success rate decreased. 

Future work can attempt to also try additional methods such as Ensemble Adversarial Training \cite{ensemble}, Adversarial Logit Pairing \cite{adversariallogitpairing}, and PGD Adversarial Training \cite{pgddefense}—with input transformation defenses like DefenseGAN \cite{defensegan}, BaRT \cite{bart}, Feature Squeezing \cite{feature} and Randomized Smoothing \cite{randomizedsmoothing} to further increase robustness against attacks.

\subsection{DL based detector}

\citet{schlarmann2024robust} proposed an unsupervised method to adversarially finetune CLIP that preserves performance on downstream tasks. We use this method for adversarial finetuning of our model. We first discuss the method below. Then we discuss our results.

\textbf{Method} We aim to minimize the distance between the embeddings of unattacked images for original CLIP model and corresponding adversarially attacked image for the robust model. The idea is that if embeddings are unaffacted by attack then performance on downstream task will be preserved. In the following we denote with $\phiorg$ the original CLIP encoder which is frozen. $\phi(z)$ is the encoder that is being trained. x is the input image and z is the corresponding adversarially attacked image. We propose the following embedding loss:

\begin{align}\label{eq:fare}
L_{\textrm{ours}}(\phi,x)=&\,\max_{\norm{z-x}_\infty \leq \epsilon} \norm{\phi(z)-\phiorg(x)}^2_2.
\end{align}

\textbf{Training Details} We use PGD attack for training with $\epsilon$ 4 and $l_\infty$ norm. We use 10 iterations of attack per image with batch size of 128. We train for 20000 iterations on ImageNet dataset. We use ADAM optimizer with a learning rate of 1e-5, weight decay of 1e-4 and learning rate warmup of 1400 steps. We use L2 loss for attack. The training parameters are consistent with \citep{schlarmann2024robust}.

\textbf{Results} We do the adversarial evaluation for just the finetuned version of our detector with DFN ViT-H backbone and 224 resolution. We evaluate using 100 iterations of APGD attack with binary cross entropy loss. We evaluate non-robust and adversarially finetuned model on clean data, attacked data with epsilon 4 and attacked data with epsilon 8. We present the results in \Cref{tab:robust_dldetector}. We observe that the adversarially trained model's performance does not degrade much on clean data but it performs much better on attacked data.

\section{Searching pHash in the Database}\label{sec:benchmark-fhew}
We have implemented our protocol as a PoC in an open-source library, available on GitHub. For this, we utilize Phantom Zone~\cite{PhantomZone}, an experimental library designed for the practical realization of MP-FHE. In our benchmarking scenario, the pHash values of the images stored in the database are encrypted using a shared key between two parties. This allows us to efficiently query and match encrypted pHash values, demonstrating the feasibility of our protocol in real-world applications.

In our implementation, we use the \texttt{FheBool} class from the PhantomZone library to perform homomorphic boolean logic. In this approach, each bit of the vectors is encrypted individually. As a result, encrypting a 96-bit vector involves creating 96 individual ciphertexts. A key advantage of treating each bit separately is the potential for parallelization during the evaluation phase of MP-FHE, which significantly speeds up the computation.

When a specific pHash value is queried, it is first encrypted locally by the client before being sent to the server for comparison with the encrypted database. The server then performs an XOR operation between the queried pHash value and each database entry. This operation is executed bit by bit and in parallel across all bits of the vectors. After the XOR operations are completed, the Hamming distance for each result is computed. This step is optimized using a 6-layer parallel boolean logic adder tree, which efficiently calculates the Hamming distance for all entries. If the distance for an entry is below the specified threshold (e.g., 8 bits), the result for that entry is set to True (1); otherwise, it is set to False (0). Once all entries are processed, the results are combined using a parallel OR tree. The final result is then sent back to the querying party for partial decryption. If the decrypted result is 1, it indicates that the queried pHash value is sufficiently similar to one or more entries in the encrypted database.

\vspace{-0.5cm}\paragraph{\textbf{Method Complexity.}}
The employed MP-FHE scheme supports Boolean operations on encrypted data. For simplicity, we assume each bit of the pHash values is encrypted individually
% \footnote{In practice, multiple bits are compressed into a single ciphertext to reduce communication and computation complexity}.
Thus, for pHash values of length $l$, we have $l$ ciphertexts. 
However, these ciphertexts are further compressed into a single one to minimize communication complexity, requiring only a few kilobytes of storage.
Our MP-FHE setup includes a parallel XOR array of length $l$ and a bit-counter of size $l$, which outputs the count of $1$s in an encrypted value. A comparator of size $\log l$ then compares this count with an encrypted threshold value, $Enc(\tau)$, and outputs an encrypted bit: it is $0$ if the Hamming distance exceeds the threshold, indicating a difference above this limit. This comparison process is repeated in parallel for every encrypted value in the registry. Finally, an OR tree with $n$ leaves calculates the OR value of all comparisons, where $n$ is the number of registry entries.
% Assuming $c$ as a constant, the overall complexity for querying one encrypted pHash against the encrypted database is: \textbf{Hamming Distance:} $O(c)$, \textbf{Comparator:} $O(\log l)$, and \textbf{OR tree:} $O(\log n)$, with the OR tree being the dominant factor, making the overall complexity equal to $O(\log n)$.

% \red{We note that as we demonstrate in Section~\ref{sec:benchmarking}, the overall computational cost of }

Table~\ref{tab:fhe_result} presents the average latency cost of determining whether two encrypted 96-bit vectors are close in FHEW, measured on a dataset with 1,000 entries. The table reports a cost breakdown of each phase of the FHEW evaluation. The columns ``XOR," ``HD," and ``Full Query" represent the average latency of calculating the XOR value, Hamming distance, and determining whether two 96-bit encrypted vectors are close or not, respectively. We conducted experiments on three different machines: 1) A midrange Dell Latitude 5531 with a 12$^{th}$ Gen Intel Core-i5 processor, 2) A Macbook Pro with a 10-core Apple Silicon M4 processor, and 3) A Google Cloud server equipped with 56 vCPUs and 112 GB memory running on an AMD EPYC 7B13 processor.

One of the key features of the proposed method is that each entry in the data is homomorphically encrypted using the same multi-party key. Consequently, FHE evaluations can be trustlessly distributed across multiple untrusted parties, as no single entity can leak or decrypt any data during query execution without access to all shared key fragments. This property enables the system to scale with minimal security considerations, making it cost-efficient. 

Although our current PoC implementation results indicate that a CPU-only device (without GPU acceleration) requires over 100 seconds to execute a query on a database with only 1,000 entries, the system can be scaled inexpensively by distributing computations across multiple trustless parties. The results from each party can then be merged using a parallel OR tree. The concept of clustering databases has also been utilized in previous work to enhance scalability~\cite{wally-search}.

\paragraph{\textbf{Related work.}} 
There are two categories of related work relevant to our constructions. The first group focuses on preserving the privacy of user queries against a database that is visible to the query executor~\cite{wally-search, tiptoe}. In such setups, it is crucial to ensure that the database owner cannot infer any information about the user's queries.

However, our system model is fundamentally different from these works. Specifically, in our target application, the database contains confidential user history that must remain undisclosed to any entity at all times. The key distinction is that, in our setting, even the data owner should not be able to reconstruct the query database. This constraint makes our protocol and the underlying problem significantly more complex than those addressed in~\cite{tiptoe, wally-search}.

Another line of work targets the same type of scenario as ours, where the database must remain unreconstructible due to its sensitive nature of stored data. The most notable recent work in this category are~\cite{iris-search, li2024panther}, in which the authors propose secretly sharing the original database among multiple parties, enabling user queries to be executed in an MPC-based manner. However, our MP-FHE-based protocol achieves stronger security guarantees. Notably:
\begin{compactitem}
    \item Query execution can be performed by any party, not just the shareholders.
    \item There are no costs associated with maintaining database shares, as the entire database remains encrypted and does not leak any information.
    \item No interaction is required among shareholders during query execution; the only interaction occurs when decrypting the final query result. This results in significantly lower communication complexity as pointed in recent work, such as~\cite{li2024panther}.
\end{compactitem}

\noindent
However, it is important to acknowledge that a pure MPC-based setup generally achieves higher performance compared to our FHE-based approach~\cite{li2024panther}. Therefore, while our construction represents an ideal and highly secure solution, it may not be practical for large-scale deployments. For this reason, can also extend and refine the MPC-based World Iris protocol~\cite{iris-search}. Other algorithms, such as KD-Trees, allow indexing for nearest-neighbor search of multi-dimensional vectors, however these frameworks cannot be used in conjunction with FHE schemes and keep perfect privacy. 

% In comparison with two TODO: double-check with phantom-zone. 

% in Tiptoe~\cite{?}, a private search engine, a single client query is 21 MB and requires 339 core-seconds of server computing against a database of 400 million entries. The high computational cost is because, for each client query, the server must scan the entire database; otherwise, it will learn which database entries the client is not interested in. Similarly, the client must send cryptographic material for each database entry, which results in high communication cost. An additional limitation of Tiptoe is that it requires each client to store a database-dependent state of around 50 MB for each client query. In wally protocol, the server computation and the communication are amortized over the number of clients making the requests; therefore, when the number of clients is large, the per-query cost becomes minimal. In Wally, the server learns a noised differentially private histogram of anonymized queries from all clients, over a partition of the server database. We emphasize that previous works, e.g., Tiptoe, are fully oblivious. That is, the server learns nothing besides that a particular client is making a query. Therefore, the privacy guarantee of schemes like Tiptoe is stronger than Wally. However, differential privacy is an accepted standard for strong privacy. 

\begin{table}[t]
\centering
\footnotesize
% \begin{threeparttable}
\begin{tabular}{cccccc}
\hline
System & CPU & Cores & XOR & HD & Full Query \\ \hline
Laptop & 12$^{th}$Gen i5 & 4 & 200 ms & 1.29 s & 1.3 s \\
Laptop & 12$^{th}$Gen i5 & 8 & 105 ms & 747 ms & 773 ms \\
Laptop & 12$^{th}$Gen i5 & 16 & 67 ms & 467 ms & 472 ms \\
MB Pro & Apple M4 & 10 & 59 ms & 421 ms &  440 ms \\
Server & AMD 7B13 & 56 & 26 ms & 134 ms & 137 ms \\ \hline
\end{tabular}
\caption{The average latency cost of determining whether two encrypted 96-bit vectors are close in FHEW (Measured on a Dataset with 1,000 entries).}
\label{tab:fhe_result}
\end{table}

\section{Preserving Database Confidentiality without FHE, using MPC}\label{sec:apx-mpc}

\autoref{fig:protocol_mpc} presents an overview of our second proposal, comprising the following key steps:

\begin{figure}[t]
  \centering
  \includegraphics[width=\linewidth,trim={0 0 3.2cm 0},clip]{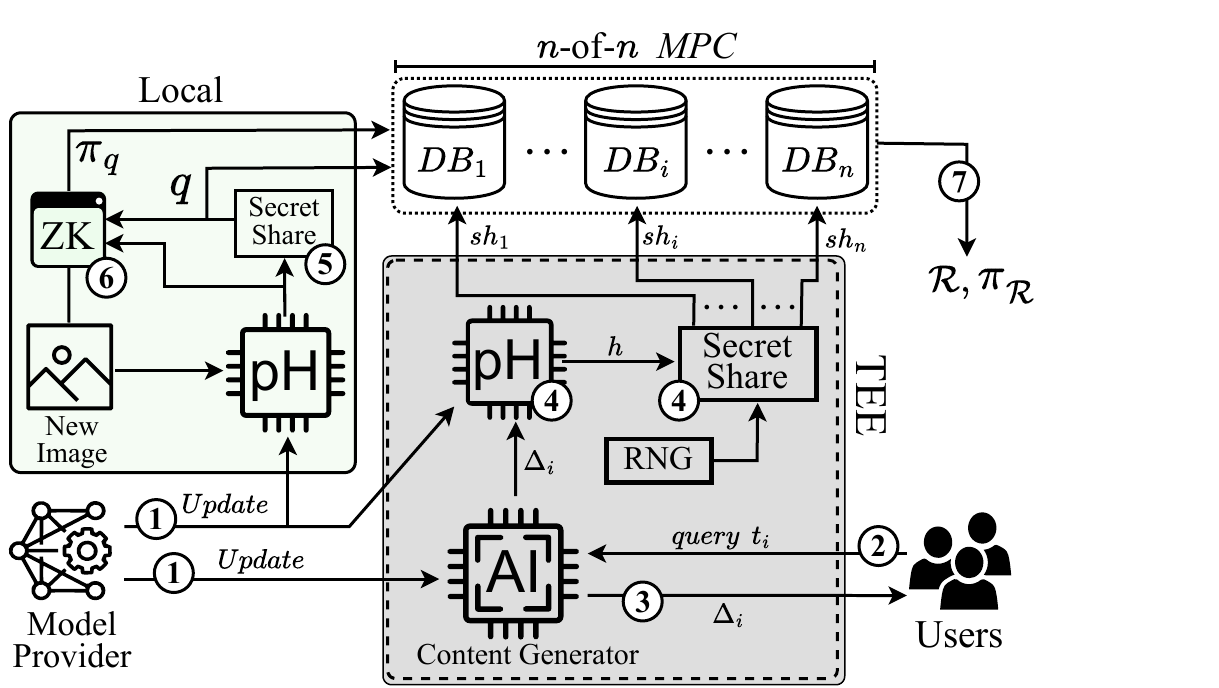}
   \caption{High level overview of the protocol.}
   \label{fig:protocol_mpc}
\end{figure}

\begin{compactitem}
    \item[] \circled{1} The \textit{model provider} is responsible for fine-tuning and updating the generative model~(e.g., OpenAI) with write-only access to the TEE. This ensures that the \textit{model provider} cannot eavesdrop on user queries. Additionally, the \textit{model provider} can update the perceptual hash function as needed.\\

    \item[] \circled{2}, \circled{3} The user $u_i$ submits a prompt to the server and receives the generated image as the result.\\
    
    \item[] \circled{4} The TEE calculates the pHash of the resulting image $\Delta_i$ and distributes the shared pHash components~($sh_1,\dots, sh_n$) among $n$ parties~($DB_1,\dots,DB_n$). During this phase, neither the generated image nor its related information is stored. No single party can retrieve any information about the image without collaboration from the others.\\
    
    \item[] \circled{5} When a new image is checked against the shared database, its pHash is first calculated, and then this value is distributed among the $n$ parties.\\
    
    \item[] \circled{6} The proof of ownership for the original image is calculated and sent to the databases alongside the shared pHash value. Given the input size of the pHash model is $224 \times 224$, the owner of the query image $x$ proves knowledge of a larger image $X$ that has been resized to $224 \times 224$. The user also demonstrates that the resized image $x$ has a pHash value of $q$, which is shared as $[q]=\{q_1,\dots,q_n\}$. Moreover, since no party will ever gain access to all of these shares~($q_1,\dots,q_n$), it is necessary to prove to each party that they posses a unique and correct share of of $[q]$. To this end, we propose building a Merkle tree $\Delta$ out of these shares and prove each party is given a certain unique leaf of $\Delta$. Finally, to connect these claims, commitments for both the original and resized versions of the image are needed, for which we employ collision-resistant image hashing techniques from prior work~\cite{vimz, veritas}. More precisely, the user asserts the following statement:
    \begin{equation}\label{eq:state-image}
        \begin{split}
        \hspace{-0.5cm}S\bigr[\mathcal{R}_\Delta, \underline{[q], i}, h_X, h_x \bigr] \: :\: \bigr\{ \exists \: X,  x\:\:\: s.t. \:\:\: x{=}f{\downarrow}(X) \\
         \mbox{ and }  h_X=\mathcal{H}_\mathcal{C}(X) \mbox{ and } h_x= \mathcal{H}_\mathcal{C}(x) \\
        \mbox{ and } q=\mathcal{H}_\mathcal{P}(x) \mbox{  and  } \exists\:\mathcal{O}(\mathcal{R}_\Delta, q_i, i\bigr)\}
        \end{split}
        % \raisetag{20.5pt}
    \end{equation}
    We note that the prover generates $n$ distinct proofs for $i \in \{1, \dots, n\}$ to ensure each party has received a unique share of $[q]$.\\
    
    \item[] \circled{7} The parties in the MPC-based database verify their received proofs and execute the query for the shared pHash value $[q]$. They then participate in the MPC to execute the given query on the shared DB. Based on the query result, the parties engage in a collaborative zkSNARK protocol to achieve public auditability, as follows:
    \begin{compactitem}
        \item \textbf{Match:} If there is a match, the result of the execution includes the index of the matched values. Therefore, the parties claim the following statement:
        \begin{equation}\label{eq:state-match}
            \begin{split}
            \hspace{-0.5cm}S\bigr[\mathcal{R}_\Delta, [q] ,t\bigr] \: :\: \bigr\{ \exists \: id\in \mathbb{N}\:\:\: s.t. \:\:\: \gamma=\prod_0^n \mathit{DB}_i[id]\\
            \hspace{-0.5cm}\mbox{ and } \mathcal{R}_\Delta=\mbox{\texttt{Merkle\_Root}}(\cup \: q_i)\mbox{ and }  \mbox{\texttt{HD}}(q,\gamma) {\leq} t\bigr\}
            \end{split}
            % \raisetag{20.5pt}
        \end{equation}
        \item \textbf{No Match:} If there are no matches found in the database, then parties will claim following statement:
        \begin{equation}\label{eq:state-no-match}
            \begin{split}
            \hspace{-0.5cm}S\bigr[\mathcal{R}_\Delta, [q] ,t\bigr] \: :\: \bigr\{ \forall \: id\in \mathbb{N}_{\{DB\}}\: \bigr|\: \gamma=\prod_0^n \mathit{DB}_i[id]\\
            \hspace{-0.5cm}\mbox{ and } \mathcal{R}_\Delta=\mbox{\texttt{Merkle\_Root}}(\cup \: q_i)\mbox{ and } \mbox{\texttt{HD}}(q,\gamma) {>} t \bigr\}
            \end{split}
            % \raisetag{20.5pt}
        \end{equation}
    \end{compactitem}
    It should be noted that while proving the statement in equation~\ref{eq:state-match} is relatively feasible, the statement in equation~\ref{eq:state-no-match} is computationally expensive. However, we argue that in most scenarios, the most critical supporting proof would be the case where a match is found, as expressed in equation~\ref{eq:state-match}.

\end{compactitem}